\documentclass[12pt]{article}
\usepackage[utf8]{inputenc}
\usepackage{
  amsmath, amssymb, amsthm, mathrsfs, calc,
  geometry, graphicx,
  txfonts, 
  ifthen, lscape,
  makeidx, manfnt,
  multicol, multirow,
  setspace, soul,
  tabto,
  textcomp,
  varioref, verbatim,
  wasysym, wrapfig,
  lipsum, eso-pic,
  eso-pic, fancyhdr,
  url, lineno,
  tikz, bm,
  subfig,
  algorithmic,tabularx
}
\usepackage[pdftex,linkcolor=blue,citecolor=blue,backref=page]{hyperref}
\usepackage[linesnumbered,lined,boxed]{algorithm2e}
\SetNlSty{}{}{}
\let\oldnl\nl
\newcommand\nonl{
  \renewcommand{\nl}{\let\nl\oldnl}}

\title{\textbf{A Comparative Study of \\ Pre-training and Self-training}} 
\author{Yiheng Wang$^*$\ \  Jiayu Lin\footnote{Equal contributions.$^{\dag}$ Correspondence authors.} $^{\dag}$ \ \ Zuoquan Lin$^{\dag}$
\\ Information and Computation Science Department
\\ Peking University\\
\texttt{wangyiheng@stu.pku.edu.cn \{linjiayu,linzuoquan\}@pku.edu.cn}
\date{}
}
   
\begin{document}
\maketitle
\sloppy

\begin{abstract}
Pre-training and self-training are two approaches to semi-supervised learning. The comparison between pre-training and self-training has been explored. However, the previous works led to confusing findings: self-training outperforms pre-training experienced on some tasks in computer vision, and contrarily, pre-training outperforms self-training experienced on some tasks in natural language processing, under certain conditions of incomparable settings. We propose, comparatively and exhaustively, an ensemble method to empirical study all feasible training paradigms combining pre-training, self-training, and fine-tuning within consistent foundational settings comparable to data augmentation. We conduct experiments on six datasets, four data augmentation, and imbalanced data for sentiment analysis and natural language inference tasks. Our findings confirm that the pre-training and fine-tuning paradigm yields the best overall performances. Moreover, self-training offers no additional benefits when combined with semi-supervised pre-training. 
\footnote{Our codes are available at https://github.com/PKUAI-LINGroup/PAS.}

%\\ \\ \textbf{Keywords} Pre-training, Self-training, Fine-tuning, Self-supervised learning, Semi-supervised learning.
\end{abstract}

\section{Introduction}

\emph{Semi-supervised learning} (SSL) involves the utilization of both labeled and unlabeled data, typically relies on a constrained amount of labeled data, and improves learning performance through the incorporation of a larger set of unlabeled data (for surveys, see \cite{van2020survey,yang2023survey}). \emph{Pre-training} and \emph{self-training} are two approaches in SSL (for surveys, see \cite{zhou2023comprehensive,amini2022self}). While pre-training and self-training share similarities that leverage unlabeled data, their methodologies and applications also have distinct differences. 

In pre-training, a model is initially trained on a large amount of unlabeled data in a self-supervised way. This pre-trained model is then fine-tuned on smaller labeled data in a supervised way for the specific tasks. \emph{Fine-tuning} is the supervised component of semi-supervised pre-training. The pre-training and fine-tuning paradigm involves training with unlabeled data and then labeled data, which can continue multiple times. Unsupervised pre-training or self-supervised pre-training refers to the pre-training conducted without subsequent fine-tuning. The pre-training and fine-tuning paradigm yields superior results for specific tasks than unsupervised pre-training. Continual pre-training refers to the pre-training and fine-tuning paradigm conducted as an additional step to continue pre-training on task-specific unlabeled data before fully supervised fine-tuning \cite{gururangan2020dont}. 

In self-training, on the other hand, the teacher model is initially trained on a small set of labeled data. The model then makes predictions on the unlabeled data, and the data points with high-confidence predictions are pseudo-labeled and added to the labeled data, resulting in the student model. The model is trained on this expanded labeled and pseudo-labeled data, and the process is iterated. The teacher and student paradigm involves training first with labeled data and then acquiring high-confidence pseudo-labels from additional unlabeled data. Self-training incorporates a form of label propagation through pseudo-labeling from unlabeled data, effectively extending the labeled data. Pre-training does not involve label propagation, instead, it centers on representation learning through patterns and structures inherent in unlabeled data. 

Due to the prominence of pre-trained large language models (LLMs), pre-training remains the best practice under scaling laws (for a survey, see \cite{zhao2023survey}). While self-trained large models have yet to emerge, self-training and its interplay with pre-training have garnered increasing research interest. 

The comparison between pre-training and self-training has been explored. 
In \cite{zoph2020rethinking} experienced in computer vision (CV), the finding was that self-training is stronger than pre-training in the following sense: Self-training performed effectively in the same setup where pre-training failed. In \cite{shi2023rethinking,shi2023don} experienced in natural language processing (NLP), the finding was that pre-training is stronger than self-training in the following sense: Continual pre-training performed better than various self-training methods. These findings led to confusion. The comparison between pre-training and self-training about these findings is somewhat unfair and lacks clarity, especially given the different settings and extra techniques involved (for detail see the \autoref{sec:related}). 

In this paper, we revisit the relationship between pre-training and self-training, while also rethinking the limitations that may prevent one from improving the performance of the other. To our knowledge, we are the first to propose an ensemble method to comparatively and exhaustively investigate all feasible training paradigms combining pre-training, self-training, and fine-tuning. In particular, we employ language models, or so-called foundation models (for a survey, see \cite{Bommasani2021OnTO})), as consistent foundational settings across all paradigms of ensemble training for downstream tasks. We employ data augmentation techniques to enhance the effectiveness of self-training. We undertake an empirical study to assess the effectiveness of the ensemble paradigms, specifically targeting six datasets, four data augmentation, and imbalanced data for sentiment analysis and natural language inference tasks in NLP. Our contributions are the findings summarized as follows: 
\begin{enumerate}
    \item[(1)] We find that semi-supervised pre-training consistently outperforms self-training and all the other training paradigms, exhibiting robust performance across varying intensities of data augmentation. 
    \item[(2)] We find that the combination of pre-training, fine-tuning, and self-training yields no benefit over the pre-training and fine-tuning paradigm. In other words, self-training offers no additional benefits when combined with semi-supervised pre-training.
    \item[(3)] We find a modest decline in pre-training performance in scenarios characterized by data imbalance; conversely, other training paradigms experienced a significant reduction in efficacy.
\end{enumerate}

\section{Related works}
\label{sec:related}

The relationship between pre-training and self-training has been examined from two perspectives: first, to evaluate the relative strengths of pre-training versus self-training; and second, to investigate how combining these two methods can mutually enhance their overall effectiveness.

\noindent\textbf{Pre-training vs. self-training}. 
As the first comparative study to challenge the prevailing paradigm of pre-training with self-training \cite{zoph2020rethinking}, this research posited that self-training is stronger than pre-training experienced in CV. Specifically, the self-training demonstrated superior performance compared to the pre-training, particularly under conditions of enhanced data augmentation and increased availability of labeled data for image recognition tasks. Notably, these experiments employed unsupervised pre-training without subsequent fine-tuning. This result contrasts with the strong baseline established by pre-trained language models. It is widely acknowledged that smaller models utilized in these experiments lack the capacity for zero-shot or few-shot learning, a capability present in LLMs \cite{brown2020language,zhao2023survey}. The substantial data and strong augmentation leveraged in the self-training are not adequately mirrored in unsupervised pre-training; thus, this comparative discrepancy renders the performance comparisons between pre-training and self-training somewhat inequitable. 

In \cite{shi2023rethinking,shi2023don}, the authors argued that pre-training is stronger than self-training experienced in NLP. Specifically, continual pre-training with or without prompt templates showed superior performance to several self-training methods for natural language understanding tasks. Compared to continual pre-training in a task-specific way, the self-training methods employed back-translation as data augmentation \cite{ott-etal-2019-fairseq}. However, a comparison still needs to be made between unsupervised pre-training used in a task-agnostic manner and self-training.

\noindent\textbf{Pre-training $\&$ self-training}. 
Two complementary can be identified in combining pre-training and self-training. One involves utilizing pre-training to enhance self-training \cite{sun-etal-2020-semi-text-classification,du-etal-2021-self,xu-etal-2021-self-pre-speech,li-etal-2021-task-adaptive,shi-etal-2023-dont}. The effectiveness of self-training is heavily dependent on the quality of the pseudo labels, underscoring the importance of a high-performing initial teacher model. In this context, the teacher model of self-training is typically initialized using pre-trained language models \cite{sun-etal-2020-semi-text-classification,xu-etal-2021-self-pre-speech}, such as BERT or RoBERTa, as demonstrated in \cite{li-etal-2021-task-adaptive,du-etal-2021-self,shi-etal-2023-dont}. This paradigm enhances model calibration and has gained traction for effectively combining self-training with pre-training, showcasing a strongly additive relationship between the two methods.

The other entails employing self-training to improve pre-training \cite{zoph2020rethinking,chen-etal-2020-self-supervised-semi-supervised,du-etal-2021-self,mi-etal-2021-self,wang-etal-2021-self-semi-speech-translation,kang-etal-2022-self-improve-pre,li-etal-2023-masked}. Self-training improved upon pre-training, demonstrating a strong additive effect \cite{zoph2020rethinking}. Self-training with strong data augmentation offered complementary advantages to unsupervised and continual pre-trained language models \cite{du-etal-2021-self,mi-etal-2021-self}. Notably, most of these experiments did not conduct a comparison with the pre-training and fine-tuning paradigm. The complementary relationship between self-training and pre-training was further explored in \cite{wang-etal-2021-self-semi-speech-translation,kang-etal-2022-self-improve-pre}. In \cite{li-etal-2023-masked}, self-training was utilized in a task-specific manner as a form of unsupervised fine-tuning, aimed at improving the performance of zero-shot learning in pre-trained models. Almost all of these self-training methods rely on strong data augmentation. Therefore, it is necessary to consider the effect of data augmentation when comparing pre-training and self-training.

Historically, self-training was first applied in NLP \cite{yarowsky1995unsupervised} (originally back \cite{scudder1965probability}). In this work, we contend that a meaningful comparison between pre-training and self-training is achievable only when utilizing consistent foundational settings, particularly language models. This is especially relevant as both NLP and CV serve as downstream tasks that can be analogized to data augmentation. It is important to exclude additional training and techniques specifically developed in prior studies to prevent incomparable settings and potentially conflicting conclusions. We aim to establish a fair comparison between pre-training and self-training within the context of language models. Embracing the pre-training and fine-tuning paradigm is crucial, as it closely mirrors the teacher-student paradigm employed in self-training. Unlike previous studies, we confirm that the pre-training and fine-tuning paradigm achieves the best overall performance, with no additional benefits from combining it with self-training.

\section{Method}
\label{sec:method}

We revisit the comparison and complementarity between pre-training and self-training, while also rethinking the limitations that may prevent one from improving the performance of the other. To this end, comparatively and exhaustively, we propose an ensemble method to study all feasible training paradigms combining pre-training and self-training within consistent foundational settings. 

\noindent\textbf{Ensemble principles}. 
When considering (unsupervised) pre-training and fine-tuning as separate processes, we identify three training components: pre-training, fine-tuning, and self-training. When combining pre-training and self-training, it is crucial to determine whether fine-tuning is included in the training protocol. It's important to recognize that not all combinations of these three components are feasible or effective for training. When designing the ensemble for these three training components, we consider the following principles: 

\begin{itemize}
    \item A training component cannot occur consecutively, as adjacent identical training components are considered the same.
    \item Pre-training can only serve as the initial component of an ensemble. If the pre-trained model is initialized during training any prior training becomes irrelevant.
    \item Self-training requires the unlabeled data in iterations and can only be performed once unless additional unlabeled data becomes available.
\end{itemize}

\noindent\textbf{Paradigms and notations}.
According to the ensemble principles, we list all feasible paradigms of ensemble training. For convenience, we use the abbreviation notations for various paradigms described in \autoref{tab:notation}. 

\begin{table*}[!t]
\centering
\begin{tabularx}{\textwidth}{l|p{8.7cm}}
\hline
\textbf{Notations}&\textbf{Description}\\
\hline
F (Fine-tuning) & Supervised training.\\
\hline
P (Pre-training) & Unsupervised pre-training.\\
\hline
S (Self-training) & Self-training.\\
\hline
PF (Pre-training$\rightarrow$Fine-tuning) & Pre-training first and then fine-tuning.\\
\hline
SF (Self-training$\rightarrow$Fine-tuning) & Fine-tuning the student model in the last iteration of self-training.\\
\hline
PS (Pre-training$\rightarrow$Self-training) & Self-training iterations commence based on pre-trained initial teacher model.\\
\hline
PSF (Pre-training$\rightarrow$Self-training\\ $\rightarrow$Fine-tuning) & Fine-tuning the student model in the last iteration of PS.\\
\hline
PFS (Pre-training$\rightarrow$Fine-tuning\\ $\rightarrow$Self-training) & Fine-tune a pre-trained model as the initial teacher model of self-training.\\
\hline
PFSF (Pre-training$\rightarrow$Fine-tuning\\ $\rightarrow$Self-training$\rightarrow$Fine-tuning) & Fine-tuning the student model in the last iteration of PFS.\\
\hline
\end{tabularx}
\caption{Notations for the paradigms.}
\label{tab:notation}
\end{table*}

We leave F along, i.e. supervised training, and P, i.e. unsupervised pre-training, as baselines that are not SSL.  
Most of the previous works \cite{sun-etal-2020-semi-text-classification,zoph2020rethinking,chen-etal-2020-self-supervised-semi-supervised,du-etal-2021-self,xu-etal-2021-self-pre-speech,li-etal-2021-task-adaptive,du-etal-2021-self,mi-etal-2021-self,wang-etal-2021-self-semi-speech-translation,kang-etal-2022-self-improve-pre} belong to PFS (see the \autoref{sec:related}).  In \cite{du-etal-2021-self}, the student model was also initialized with the pre-trained model, which can be viewed as a variant of PFS (an analysis see \autoref{fig:PFS} in the \autoref{sec:experiments}). In \cite{li-etal-2023-masked}, an unsupervised classifier included self-training is similar to PS. 

\begin{figure*}
	\centering
	\includegraphics[width=0.8\textwidth]{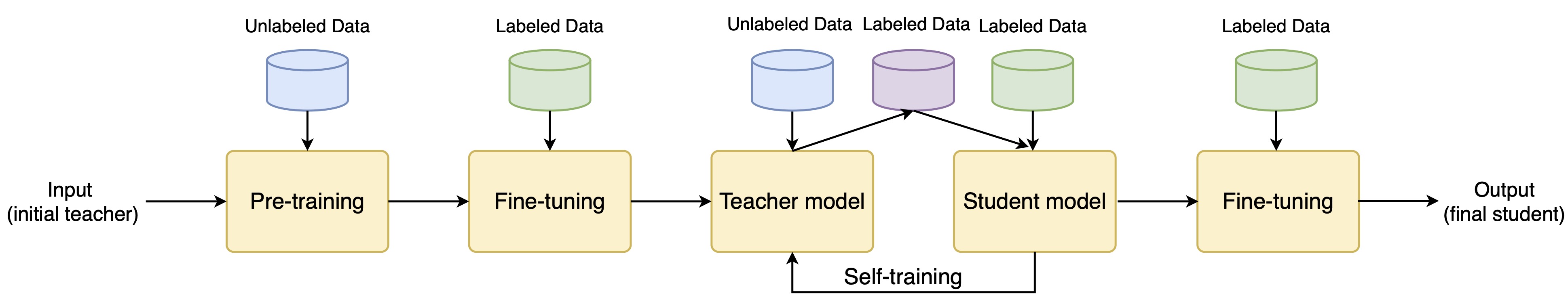}
	\caption{The training process of PFSF: Fine-tuning the student model in the last iteration of the self-training that fine-tunes a pre-trained model as its initial teacher
model.}
	\label{fig:pfsf}
\end{figure*}

SF, PSF, and PFSF have not been explored in prior research. We examine these paradigms for some considerations. One major challenge in self-training is semantic drift, where accumulating incorrect pseudo labels can misguide the training process over time. A potential solution to this problem is to fine-tune the final student model using labeled data. The complex PFSF is depicted in \autoref{fig:pfsf} (for self-training strategy refer to the explanation below). To some extent, other paradigms can be regarded as special parts of PFSF. As we shall see later, the limitation of PFSF is that increasing training costs does not necessarily bring efficiency. 

\noindent\textbf{Self-training}. 
We use a competitive version of pseudo-labeling by using a self-paced curriculum strategy in the context of self-training \cite{cascante-bonilla-etal-2021-curriculum-labeling}. Pseudo-labeling is trained incrementally by iteratively propagating labels from labeled data to unlabeled data using the model, re-labeling high-confidence predictions, and retraining with labeled and pseudo-labeled data. Instead of adding all pseudo-labeled data in each iteration in original pseudo-labeling \cite{lee2013pseudo}, self-pace pseudo-labeling carefully selects a subset of the most confident data to help guide the model towards harder samples in a controlled manner, improving performance. The algorithm is briefly described as follows:

\begin{enumerate}
    \item[(1)]  \emph{Train}: The teacher model is first trained on the labeled data.
    \item[(2)] \emph{Predict}: Pseudo-labels are assigned to the unlabeled data using the current model.
    \item[(3)] \emph{Select}: A subset of pseudo-labeled data is selected based on their prediction scores and percentile thresholds.
    \item[(4)]  \emph{Re-train}: The student model is trained from scratch using both labeled and selected pseudo-labeled data.
    \item[(5)] \emph{Repeat}: Steps (2-4) are repeated until all data in the dataset have been used during training.
\end{enumerate}

\noindent To alleviate concept drift and confirmation bias, the model parameters are reinitialized before each iteration. This ensures that previous erroneous predictions do not accumulate over time (for detail, refer to \cite{cascante-bonilla-etal-2021-curriculum-labeling}).

\noindent\textbf{Language models}. 
We employ language models as consistent foundational settings across all paradigms of ensemble training for various downstream tasks. Specifically, we utilize the transformer-based BERT model as our initial backbone \cite{devlin-etal-2019-bert}. 

We value BERT's encoding representation capability, as we do not primarily consider the generation capability of language models. Moreover, we choose the basic BERT model by two key considerations: first, we aim to avoid using stronger pre-trained language models to maintain a level playing field for self-training; second, we know that utilizing pre-trained language models as the initial teacher model enhances the self-training process. The effectiveness of self-training is heavily dependent on the calibration of the teacher model, as inaccurate pseudo-labels generated by the initial teacher can misguide the training of the student model.

\noindent\textbf{Data augmentation}.
Data augmentation (DA) artificially increases the size of a training dataset by generating modified versions of existing data points, addressing the challenge of limited labeled data like SSL. Previous research has shown that experiments favoring self-training over pre-training often employed data augmentation to enhance the effectiveness of self-training. Consequently, we investigate the impact of four data augmentation strategies of varying intensities on different paradigms of ensemble training, including natural noise, conditional BERT, and back-translation.

Natural noise is a data augmentation technique in NLP that simulates common human errors, introducing character-level and word-level mistakes to enhance comprehension  \cite{belinkov-etal-2018-noise-break-translation}. Conditional BERT addresses data-label mismatch via masked language modeling, allowing it to generate sentences aligned with specific labels during fine-tuning \cite{wu-etal-2019-bert-augment}. Additionally, back-translation involves translating text to a target language and back to the source to create augmented data that retains the original meaning while varying its form \cite{sennrich-etal-2016-improving}, facilitated by tools like Fairseq \cite{ott-etal-2019-fairseq}.

\section{Experiments}
\label{sec:experiments}

\subsection{Datasets}

We conduct experiments on two tasks in NLP: sentiment analysis (SA) and natural language inference (NLI). NA identifies the emotions and feelings expressed in text and is a text classification problem with two or more classes. We use four datasets: IMDB \cite{maas-etal-2011-learning}, SST \cite{socher-etal-2013-recursive}, AG News \cite{zhang-etal-2015-character-cnn} and Elec \cite{singh-etal-2019-elec}. NLI judges whether the premise and the hypothesis match, and the result can be \emph{True}, \emph{False}, and \emph{Undetermined}. We use two datasets: SNLI \cite{bowman-etal-2015-large} and MultiNLI \cite{williams-etal-2018-broad}. The statistics of the datasets are shown in \autoref{tab:datasets}.

\begin{table}[!t]
	\centering
	\begin{tabular}{|c|c|c|c|c|}
		\hline
		\textbf{Datasets}&\textbf{Labeled}&\textbf{Unlabeled}&\textbf{Valiadation}&\textbf{Test}\\
		\hline
		IMDB&15,000&50,000&25,000&25,000\\
\hline
		SST&7,349&60,000&1,800&1,800\\
\hline
		AG News&10,000&50,000&7k600&7,600\\
\hline
		Elec&25,000&200,000&25,000&25,000\\
\hline
		SNLI&10,000&50,000&10,000&10,000\\
\hline
		MultiNLI&10,000&50,000&10,000&10,000\\
		\hline
	\end{tabular}
        \caption{Statistics of the datasets.}
	\label{tab:datasets}
\end{table}

\subsection{Implementations}

We employ BERT to map input text into a feature space. We attach a linear layer as a classifier atop the BERT model for classification tasks. We utilize BERT in two configurations: BERT-medium, which comprises 8 layers with a hidden size of 512, 8 attention heads, and an intermediate size of 2048 \cite{bhargava-etal-2021-generalization, turc-etal-2019-student-init}, and BERT-base, which comprises 12 layers with a hidden size of 768, 12 attention heads, and an intermediate size of 3072 \cite{devlin-etal-2019-bert}.

For the select step in self-training (refer to the \autoref{sec:method}), we retrieve the top $R\%$ (the multiples of 10 or 20) confident data with $R$ improving as the number of iterations increases from all unlabeled data. As usual, we set the learning rate as 1e-5 and batch size as 64 and trained the model within 20 epochs and 40 epochs for BERT-medium and BERT-base respectively.

\subsection{Results}

We conduct experiments for each paradigm of ensemble training on all the datasets to observe the performance. The results are shown in \autoref{tab:accuracy}, from which we can find the following facts:

\begin{itemize}
    \item Self-training (S) is effective, surpassing the baselines (F and P).
    \item The pre-training and fine-tuning paradigm (PF) demonstrates the best performance across all the datasets. This verifies the superiority of PF. 
    \item The accuracy of S, PS, and PFS are close, which reveals the invalidity of the pre-trained teacher model with or without fine-tuning (see more discussions in the \autoref{sec:discussions}). 
    \item Fine-tuning has either resulted in negligible improvement or a slight decline in the performance of S, PS, and PFS, which indicates that the information in labeled data has already been exploited sufficiently.     
\end{itemize}

\begin{table}[!t]
	\centering	
	\begin{tabular}{|c|cccc|cc|}
		\hline
		\multirow{2}{*}{\textbf{Paradigms}}&\multicolumn{4}{c|}{\textbf{NA}}&\multicolumn{2}{c|}{\textbf{NLI}}\\
		\cline{2-7}
		&IMDB&SST&Elec&AG News&SNLI&MultiNLI\\
		\hline
		F&.8391&.7580&.8775&.8453&.5274&.4343\\
\hline
		P&.5000&.5000&.5000&.2500&.3333&.3333\\
\hline
		S&.8556&.7695&.8776&.8800&.5427&.4424\\
\hline
PF&\textbf{.8878}&\textbf{.8658}&\textbf{.9246}&\textbf{.8882}&\textbf{.7696}&\textbf{.6578}\\
\hline
		SF&.8404&.7661&.8747&.8739&.5407&.4448\\
\hline		
		PS&.8540&.7833&.8766&.8803&.4703&.4245\\
\hline
		PSF&.8451&.7775&.8769&.8672&.5306&.4290\\
\hline
		PFS&.8551&.7672&.8778&.8797&.5344&.4502\\
\hline
		PFSF&.8482&.7649&.8762&.8786&.5408&.4426\\
		\hline
	\end{tabular}
        \caption{Accuracy of the paradigms.}
        \label{tab:accuracy}
 \end{table}
 
\subsection{Data augmentation}

We perform experiments to assess the effectiveness of varying intensities of data augmentation within ensemble paradigms. We create four data augmentation strategies by integrating natural noise, conditional BERT, and back-translation to ensure increased data augmentation. These strategies are designated as DA1, DA2, DA3, and DA4, as detailed in \autoref{tab:augmentation}, where we write DA0 for no data augmentation for the sake of comparison.

\begin{table}[!t]
	\centering	
	\begin{tabular}{|c|l|}
		\hline
		\textbf{Strategy} & \textbf{Description}\\
            \hline
		DA0 & No data augmentation.\\
		\hline
		DA1 & Natural noise.\\
            \hline
		DA2 & Conditional BERT and natural noise.\\
            \hline
		DA3 & Back-translation, conditional BERT, and natural noise.\\
            \hline
		DA4 & The same as DA3 with larger magnitude.\\
		\hline
	\end{tabular}
        \caption{Data augmentation strategies.}
        \label{tab:augmentation} 
\end{table}

We perform experiments using two datasets: IMDB and SST. We begin by sampling 1,000 instances evenly from each class as labeled data while leaving the unlabeled data unchanged. We then augment the labeled data to a total of 10,000 instances. The objective is to investigate the effects of pre-training intensity and the degree of data augmentation. The findings are detailed in \autoref{tab:bert-medium-imdb}, \autoref{tab:bert-base-imdb}, \autoref{tab:bert-medium-sst}, and \autoref{tab:bert-base-sst}.  

\begin{table*}[!t]
	\centering	
	\begin{tabular}{|c|ccccc|}
		\hline
		\textbf{Paradigms}&\textbf{DA0}&\textbf{DA1}&\textbf{DA2}&\textbf{DA3}&\textbf{DA4}\\
		\hline		F&.5641&.7680(+20.39\%)&.7697(+20.56\%)&.7748(+21.07\%)&.7741(+21.00\%)\\
\hline		S&.7595&.8199(+6.04\%)&.8197(+6.02\%)&.8218(+6.23\%)&.8178(+5.83\%)\\
\hline		SF&.7888&.8132(+2.44\%)&.8013(+1.25\%)&.7958(+0.7\%)&.8168(+2.80\%)\\
\hline
		PT&\textbf{.8393}&\textbf{.8413}(+0.02\%)&\textbf{.8411}(+0.18\%)&\textbf{.8446}(+0.53\%)&\textbf{.8377}(-0.16\%)\\
\hline		PS&.7370&.8134(+7.64\%)&.8104(+7.34\%)&.8170(+8\%)&.8120(+7.5\%)\\
\hline		PSF&.7798&.8021(+2.23\%)&.8017(+2.19\%)&.8095(+2.97\%)&.8136(+3.38\%)\\
\hline		PFS&.7752&.8201(+4.49\%)&.8176(+4.24\%)&.8191(+4.39\%)&.8192(+4.40\%)\\
\hline		PFSF&.7908&.8141(+2.33\%)&.8090(+1.82\%)&.8066(+1.58\%)&.8202(+2.94\%)\\
		\hline
	\end{tabular}
        \caption{Accuracy of the paradigms on IMDB using BERT-medium.}
        \label{tab:bert-medium-imdb} 
\end{table*} 

\begin{table*}[!t]
	\centering	
	\begin{tabular}{|c|ccccc|}
		\hline		\textbf{Paradigms}&\textbf{DA0}&\textbf{DA1}&\textbf{DA2}&\textbf{DA3}&\textbf{DA4}\\
		\hline		F&.5001&.7713(+27.12\%)&.7742(+27.41\%)&.7685(+26.84\%)&.7662(+26.61\%)\\
\hline		S&.7927&.8202(+2.75\%)&.8162(+2.35\%)&.8209(+2.82\%)&.7794(-1.33\%)\\
\hline		SF&.8023&.8195(+1.72\%)&.8164(+1.41\%)&.8180(+1.57\%)&.7867(-1.56\%)\\
\hline
		PT&\textbf{.8470}&\textbf{.8709}(+2.39\%)&\textbf{.8742}(+2.72\%)&\textbf{.8807}(+3.37\%)&\textbf{.8736}(+2.66\%)\\
\hline		PS&.7741&.8222(+4.81\%)&.8179(+4.38\%)&.8252(+5.11\%)&.7967(+2.26\%)\\
\hline		PSF&.7747&.8232(+4.85\%)&.8172(+4.25\%)&.8195(+4.48\%)&.8012(+2.65\%)\\
\hline		PFS&.7827&.8182(+3.55\%)&.8127(+3.00\%)&.8182(+3.25\%)&.8074(+2.47\%)\\
\hline		PFSF&.7830&.8195(+3.65\%)&.8157(+3.27\%)&.8180(+3.50\%)&.8117(+2.87\%)\\
		\hline
	\end{tabular}
        \caption{Accuracy of the paradigms on IMDB using BERT-base.}
        \label{tab:bert-base-imdb} 
\end{table*} 

\begin{table*}[!t]
	\centering	
	\begin{tabular}{|c|ccccc|}
		\hline		\textbf{Paradigms}&\textbf{DA0}&\textbf{DA1}&\textbf{DA2}&\textbf{DA3}&\textbf{DA4}\\
		\hline		F&.5722&.7007(+12.85\%)&.6984(+12.62\%)&.6972(+12.5\%)&.6961(+12.39\%)\\
\hline		S&.6927&.7259(3.32\%)&.7133(+2.06\%)&.7087(+1.6\%)&.7087(+1.6\%)\\
\hline		SF&.6869&.7236(+3.67\%)&.7167(+2.98\%)&.6915(+0.46\%)&.7156(+2.87\%)\\
\hline
		PT&\textbf{.8096}&\textbf{.8257}(+1.61\%)&\textbf{.8119}(+0.23\%)&\textbf{.7982}(-1.14\%)&\textbf{.8073}(-0.23\%)\\
\hline		PS&.5401&.6720(+13.19\%)&.6307(+9.06\%)&.6984(+15.83\%)&.6800(+13.99\%)\\
\hline		PSF&.5849&.6755(+9.06\%)&.6755(+9.06\%)&.6892(+10.43\%)&.6915(+10.66\%)\\
\hline		PFS&.6697&.7213(+5.16\%)&.7236(+5.39\%)&.7133(+4.36\%)&.7259(+5.62\%)\\
\hline		PFSF&.6846&.7053(+2.07\%)&.7179(+3.33\%)&.6915(+0.69\%)&.7144(+2.98\%)\\
		\hline
	\end{tabular}
        \caption{Accuracy of the paradigm on SST using BERT-medium.}
        \label{tab:bert-medium-sst} 
\end{table*}

\begin{table*}[!t]
	\centering	
	\begin{tabular}{|c|ccccc|}
		\hline		\textbf{Paradigms}&\textbf{DA0}&\textbf{DA1}&\textbf{DA2}&\textbf{DA3}&\textbf{DA4}\\
		\hline		F&.5092&.6881(+17.89\%)&.6846(+17.54\%)&.6991(+18.99\%)&.6778(+16.86\%)\\
\hline		S&.6250&.7385(+11.35\%)&.7179(+9.29\%)&.7213(+9.63\%)&.7397(+11.47\%)\\
\hline		SF&.6904&.7351(+4.47\%)&.7122(+2.18\%)&.7156(+2.52\%)&.7339(+4.35\%)\\
\hline
		PT&\textbf{.8773}&\textbf{.8865}(+0.92\%)&\textbf{.8716}(-0.57\%)&\textbf{.8784}(+0.11\%)&\textbf{.8693}(-0.8\%)\\
\hline		PS&.5952&.7076(+11.24\%)&.6709(+7.57\%)&.7167(+12.15\%)&.6479(+5.27\%)\\
\hline		PSF&.5688&.7099(+14.11\%)&.6686(+9.98\%)&.7053(+13.65\%)&.6823(+11.35\%)\\
\hline		PFS&.6892&.7420(+5.58\%)&.7305(+4.43\%)&.7225(+3.63\%)&.7351(+4.89\%)\\
\hline		PFSF&.6823&.7443(+6.2\%)&.7259(+4.36\%)&.7202(+3.79\%)&.7305(+4.82\%)\\
		\hline
	\end{tabular}
        \caption{Accuracy of the paradigms on SST using BERT-medium.}
        \label{tab:bert-base-sst} 
\end{table*}

We've omitted the accuracy of P in these tables due to its trivial nature. The number in the bracket indicates the change magnitude relative to DA0. We find two trends regarding accuracy as the magnitude of data augmentation increases:

\begin{itemize}
    \item Accuracy initially rises and then declines as the extent of the data augmentation strategy grows.
    \item Accuracy increases initially and then stabilizes, indicating that moderate data augmentation enhances performance, whereas excessive augmentation is ineffective and may even hinder results.
\end{itemize}

Additionally, the pre-training and fine-tuning paradigm demonstrates greater stability than other paradigms. It shows resistance to variations in data augmentation magnitude and does not depend on sufficient labeled data.

Our observations reveal that when using pre-trained weights, stronger pre-training knowledge (BERT-base) outperforms weaker pre-training knowledge (BERT-medium) in scenarios with no data augmentation and moderate data augmentation. While most paradigms that utilize pre-training knowledge, apart from PF, struggle with strong pre-training under excessive data augmentation, PF shows improvement. Specifically, PS, PFS, PSF, and PFSF exhibit poorer performance or only marginal increases compared to PF. Employing a stronger pre-training model results in a wider performance gap between PF and the other paradigms.

\subsection{Imbalanced data}

Data imbalance is a prevalent issue that hinders model performance, prompting us to examine the effectiveness of the paradigms in this context. To create imbalanced training data, we sample from the original datasets and conduct experiments on two datasets: IMDB for binary classification and AG News for four-category classification. The data ratio for IMDB is set at 1:5, while for AG News, the ratio is 1:1:1:7. The results of these experiments are demonstrated in \autoref{tab:imbalance}. The number in the bracket indicates the change magnitude relative to balanced data.

\begin{table*}[!t]
	\centering
	\begin{tabular}{|c|cc|cc|}
		\hline
		\multirow{2}{*}{\textbf{Paradigms}}&\multicolumn{2}{c}{\textbf{IMDB}}&\multicolumn{2}{c|}{\textbf{AG News}}\\		&\textit{balanced}&\textit{imbalanced}&\textit{balanced}&\textit{imbalanced}\\
		\hline
		F&.8391&.7894(-4.97\%)&.8453&.7418(-9.73\%)\\
\hline
		S&.8556&.8531(-0.25\%)&.8800&.4487(-43.14\%)\\
\hline
		SF&.8404&.8163(-2.41\%)&.8739&.4654(-40.85\%)\\	
\hline
  PF&\textbf{.8878}&.8471(-4.07\%)&\textbf{.8882}&\textbf{.8391}(-4.91\%)\\
\hline
		PS&.8540&.8532(-0.08\%)&.8803&.6566(-22.37\%)\\
\hline
		PSF&.8451&.8231(-2.2\%)&.8672&.6513(-21.59\%)\\
\hline
		PFS&.8551&\textbf{.8547}(-0.4\%)&.8797&.5633(-31.64\%)\\
\hline
		PFSF&.8482&.8042(-4.4\%)&.8786&.5691(-30.95\%)\\
		\hline
	\end{tabular}
        \caption{Accuracy of the paradigms on imbalanced data.}
        \label{tab:imbalance} 
\end{table*}

In binary classification, the performance of the paradigms does not experience a significant decline. However, the four-category classification shows a marked drop in performance across the paradigms. Notably, PS and PSF demonstrate greater resilience to the adverse effects of data imbalance compared to PFS and PFSF, while PFS and PFSF maintain more stability than S and SF. Although most other paradigms face substantial decreases, PF maintains consistent performance with a few value changes.

\section{Discussions}
\label{sec:discussions}

We discuss the reasons behind the failure of PFS in the experiments. Unlike previous studies, we find that self-training and pre-training do not function as complementarity. Notably, there are instances where PFS performs worse than S. To delve deeper into this issue, we analyze the evaluation accuracy for each iteration in PFS. 

\begin{figure*}[!t]
	\centering
	\subfloat[PFS Random-init on IMDB.]{\includegraphics[width=0.35\textwidth]{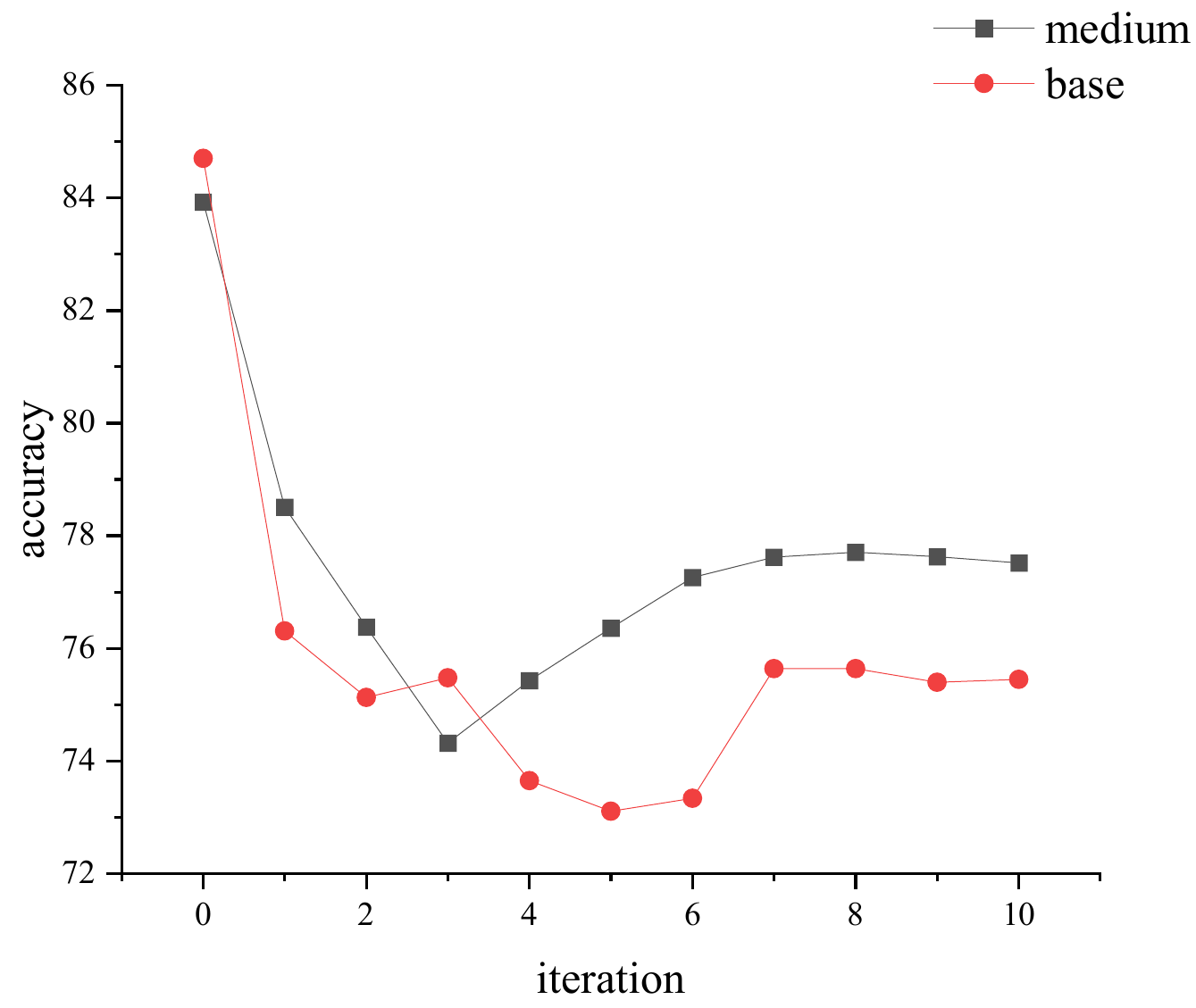}
		\label{fig3(a)}}
	\subfloat[PFS Random-init on SST.]{\includegraphics[width=0.35\textwidth]{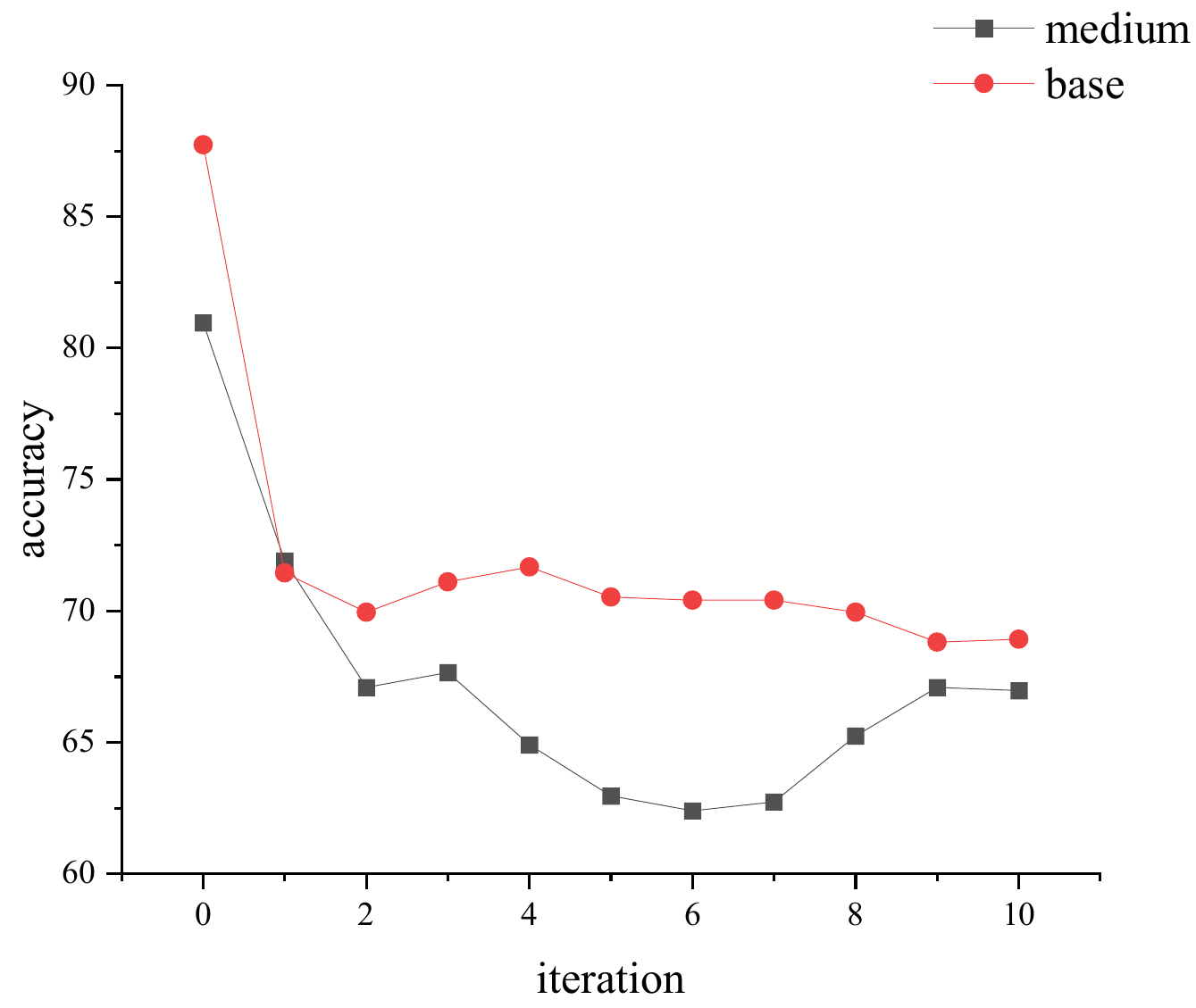}
		\label{fig3(b)}} \\
	\subfloat[PFS Pre-init on IMDB.]{\includegraphics[width=0.35\textwidth]{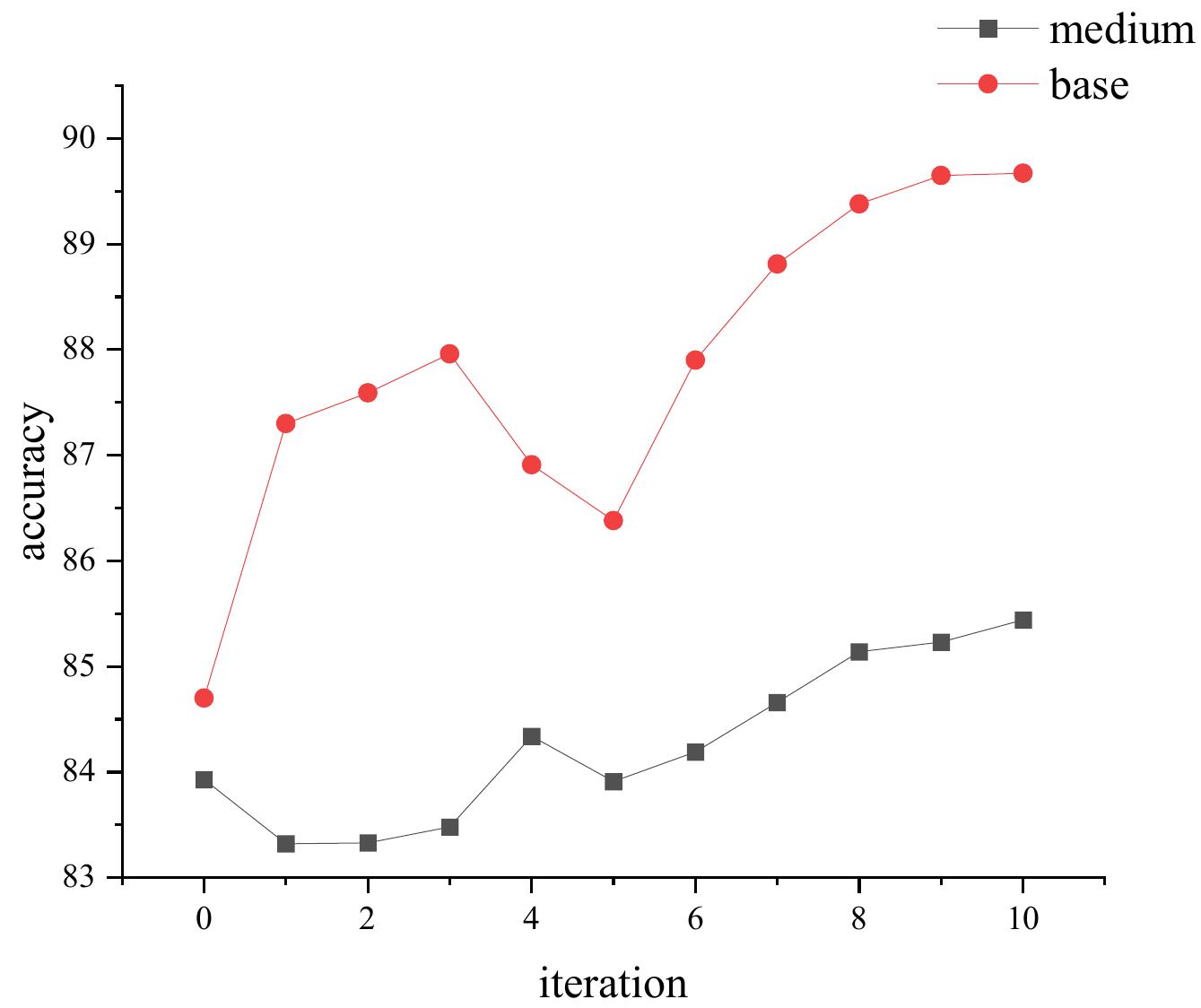}
		\label{fig3(c)}}
	\subfloat[PFS Pre-init on SST.]{\includegraphics[width=0.35\textwidth]{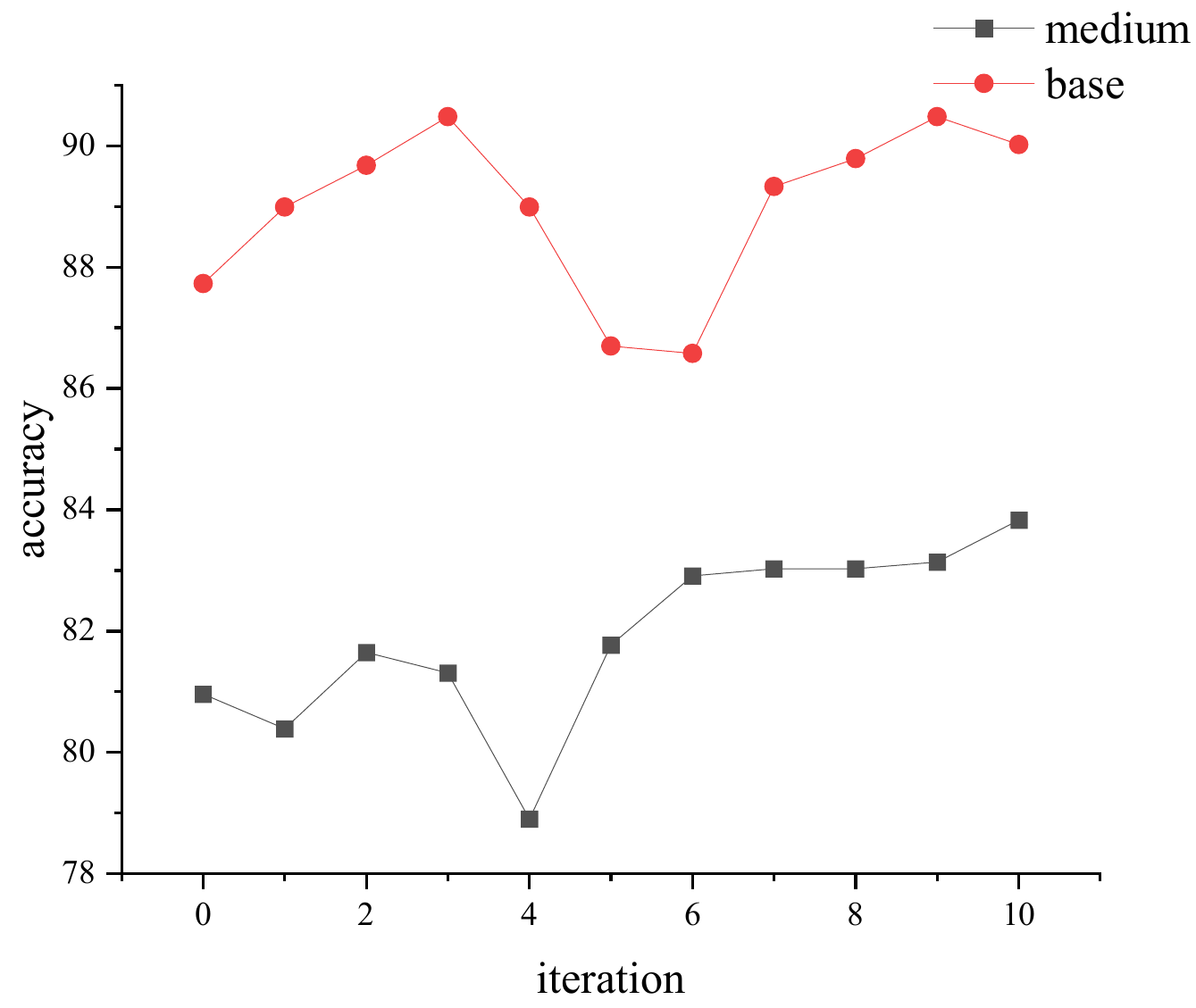}
		\label{fig3(d)}}
	\caption{Accuracy of each iteration for PFS with different initialization.}
	\label{fig:PFS}
\end{figure*}

As illustrated in \autoref{fig:PFS} (a)(b) for PFS with random initialization (written as PFS Random-init), i.e. regular self-training described in the \autoref{sec:method}, there is a significant drop in performance during the first iteration, followed by gradual improvements in subsequent iterations, but converges to poor performance. This pattern indicates that the student model in the initial iteration struggles to retain the collective knowledge gained during pre-training. We hypothesize that this may result from inefficient knowledge transfer from the pre-trained teacher model to the student model through pseudo-labels in the process of PFS. 

We consider the PFS with a student model with sufficient pre-training knowledge to test our hypothesis. To inject pre-trained knowledge into a student model, the student model is initialized with pre-trained parameters and then fine-tuned by labeled and pseudo-labeled data in each iteration (written as PFS Pre-init). We find that the PFS outperforms both PF and S, as depicted in \autoref{fig:PFS} (c)(d). 
This observation illustrates that the PFS with sufficient pre-training knowledge succeeds in improving upon PF and provides evidence to support our hypothesis.

\section{Conclusions}

We proposed an ensemble method to empirically explore all feasible training paradigms that combine pre-training, self-training, and fine-tuning with language models. Our study revisited the relationship between pre-training and self-training, while critically examining the limitations that may hinder improvements in either approach.

Our findings indicated that the pre-training and fine-tuning paradigm is the most effective among the various training paradigms. While this is not a discovery, it clarifies existing research on self-training and its interaction with pre-training. This analysis provides valuable insights for future design considerations and assists in selecting the most appropriate learning strategies.

% \section{Limitations}

% Due to the lack of self-trained large models, we cannot compare pre-training and self-training in the context of LLMs. We did not conduct experiments on CV or other NLP tasks, although we think it suffices for some NLP tasks to show the comparative study. These limitations can be further questions for consideration. 

%\section*{Acknowledgements}
%This work was supported by the National Natural Science Foundation of China under grant number 62076009.

\bibliographystyle{plain}
\bibliography{reference}

\end{document}